\pdfoutput=1

\documentclass[11pt]{article}

\usepackage[preprint]{acl}

\usepackage{times}
\usepackage{latexsym}

\usepackage[T1]{fontenc}

\usepackage[utf8]{inputenc}

\usepackage{microtype}

\usepackage{inconsolata}

\usepackage{graphicx}

\usepackage[utf8]{inputenc}
\usepackage{booktabs}
\usepackage{tabularx}
\usepackage{enumitem}
\usepackage{array}
\usepackage{amsmath}
\usepackage{multirow}

\title{TravelBench : Exploring LLM Performance in Low-Resource Domains}

\author{Srinivas Billa \\
  Expedia Group \\
  The Angel Building, 407 St John St \\ 
  London, EC1V 4EX, UK \\
  \texttt{nbilla@expediagroup.com} \\\And
  Xiaonan Jing \\
  Expedia Group \\
  1111 Expedia Group Wy W \\ 
  Seattle, WA 98119, USA \\
  \texttt{xijing@expediagroup.com} \\}

\begin{document}
\maketitle
\begin{abstract}
Results on existing LLM benchmarks capture little information over the model capabilities in low-resource tasks, making it difficult to develop effective solutions in these domains. To address these challenges, we curated 14 travel-domain datasets spanning 7 common NLP tasks using anonymised data from real-world scenarios, and analysed the performance across LLMs. We report on the accuracy, scaling behaviour, and reasoning capabilities of LLMs in a variety of tasks. Our results confirm that general benchmarking results are insufficient for understanding model performance in low-resource tasks. Despite the amount of training FLOPs, out-of-the-box LLMs hit performance bottlenecks in complex, domain-specific scenarios. Furthermore, reasoning provides a more significant boost for smaller LLMs by making the model a better judge on certain tasks.
\end{abstract}

\section{Introduction}
The rapid advancement of large language models (LLMs) in recent years has significantly facilitated the prototyping of downstream natural language processing (NLP) tasks. However, this has also introduced new challenges in selecting the most suitable LLM from an ever-growing pool of state-of-the-art (SOTA) models. To address this issue, a number of benchmarking datasets are proposed to evaluate the general capabilities of LLMs, including but not limited to: MMLU \cite{hendrycks2020measuring}, ARC-C \cite{clark2018think}, GSM8K \cite{cobbe2021training}, HumanEval \cite{chen2021evaluating}, MGSM \cite{shi2022language}. However, these benchmarks provide limited insight into the performance of the model in low-resource domains such as the travel industry. In addition, prior study \cite{wiher2022decoding} has shown that different types of generation can affect the performance of an LLM's decoding strategy. Without in-depth domain evaluation, streamlining LLM development becomes more difficult.

Traditionally, \textbf{opinion mining} has been the most prominent task in the travel domain, as it was often studied over customer reviews. For instance, hotel reviews have been widely used in sentiment classification \cite{alam2016joint}, rating prediction \cite{wang2010latent}, and opinion spam detection \cite{ott2011finding}. Similarly, restaurant reviews have been long leveraged for benchmarking the overall and aspect-based sentiment analysis, particularly in SemEval-2014 Task 4 \cite{cca5dae19d614c7c92c9acabc0428ce6}. Most recently, with the emergence of LLM agents, \textbf{trip planning} has gained attention and benchmark such as TravelPlanner \cite{xie2024travelplanner} has been introduced, which evaluates LLMs on tool usage and complex planning. To further understand LLM performance in the travel domain beyond opinion mining and broaden the scope of real-world usage scenario benchmarking, we curated a comprehensive set of travel datasets including 14 tasks across 7 categories for model evaluation. 
An overview of the datasets is presented in Table \ref{tab:nlp_tasks}. 

By applying LLMs to generate predictions, we treat each task as solving an autoregressive modelling problem. The prediction then reflects the next token generation capability given specific contexts. Thus, the output is characterized by a combination of the model's internal knowledge and instruction-following ability.
\begin{table*}
  \centering
    \begin{tabular}{llll} \hline
    \textbf{Task}                                   & \textbf{\# Samples}                  & \textbf{Avg Input Tokens}           & \textbf{\# Labels}             \\ \hline
    Aspect Based Sentiment Analysis        & \multirow{12}{*}{$\sim$500} & $\sim$700                  & 6                     \\
    Overall Review Sentiment Analysis      &                             & $\sim$300                  & 3                     \\ \cline{1-1} \cline{3-4} 
    Aspect Based Review Segmentation       &                             & $\sim$900                  & Open Ended Generation \\
    Review Topic Classification            &                             & $\sim$500                  & 17                    \\ \cline{1-1} \cline{3-4} 
    Faithfulness                          &                             & $\sim$1500                 & 5                     \\
    Relevance                              &                             & $\sim$700                  & 5                     \\
    Inclusiveness                          &                             & $\sim$1000                 & 2                     \\
    Compliance                             &                             & $\sim$900                  & 2                     \\ \cline{1-1} \cline{3-4} 
    Review Moderation                      &                             & $\sim$2000                 & 2                     \\
    Manager Response Moderation            &                             & $\sim$1400                 & 2                     \\ \cline{1-1} \cline{3-4} 
    Customer Service Intent Prediction     &                             & $\sim$800                  & 9                     \\ \cline{1-1} \cline{3-4} 
    Review Summarisation                   &                             & $\sim$2500                 & Open Ended Generation \\ \hline
    Review Translation EN -\textgreater XX & \multirow{2}{*}{$\sim$1000} & \multirow{2}{*}{$\sim$100} & Open Ended Generation \\
    Review Translation XX -\textgreater EN &                             &                            & Open Ended Generation \\ \hline
    \end{tabular}
  \caption{
    Benchmark datasets categorised by NLP task. Each task is treated as a generation problem using an auto-regressive LLM. Avg Input Tokens denotes the number of tokens in the prompt which consists both instruction tokens and data tokens. For classification problems, number of labels are reported.
  }
  \label{tab:nlp_tasks}
\end{table*}
In the rest of this paper, we present in-depth evaluations of LLMs on our benchmark, aiming to bridge the gap between \textbf{low-resource, under-explored tasks} in the travel domain and the demand of real-world applications. It should be noted that all of our datasets were collected in an anonymised form from real-world usage scenarios, making the evaluation results more representative of actual model performance in practice. Our contributions are as follows:
\begin{itemize}[nolistsep]
    \item We introduce a wide range of curated datasets in the travel domain for LLM evaluation, expanding the scope of current resources beyond opinion mining. 
    \item We present an in-depth analysis of various LLMs, verifying that out-of-the-box LLMs have performance limitations when adapted to travel domain. 
    \item We share insights on LLM's scaling and reasoning behaviours to shed light on the effects of model's training FLOPs and size.  \footnote{We are in the process of approval to release the datasets for further research.}
\end{itemize}

\section{Dataset}
In this section, we describe the datasets, data collection process, and finally the evaluation metrics. The selection of tasks was derived from real-world downstream tasks in the travel domain. We sourced data from real-world scenarios, i.e. publicly deployed systems. The data was anonymised to remove Personally Identifiable Information (PII). All of the data was human annotated or verified without assistance from LLMs, minimizing implicit bias in the dataset towards models. For instance, we omitted sample responses to summary generation in the annotation guideline to reduce the impact of writing styles from pre-defined summaries. Dataset curation follows three steps:
\begin{enumerate}[nolistsep]
    \item \textbf{Stratified Random Sampling.} From a large collection of real-world usage, we randomly sample approximately 500 to 1000  rows per task following the source distribution.
    \item \textbf{Rubric Creation.} We developed annotation guidelines in collaboration with human experts to provide precise instructions and ensure label consistency.
    \item \textbf{Annotation.} We employed human coders for annotation. It should be noted that although the number of coders could vary across datasets, the guidelines were designed to minimize variance introduced by these differences.
\end{enumerate}

\subsection{Aspect-based Sentiment Analysis (ABSA)}

The purpose of this task is to identify granular sentiment toward specific topics within a text. Given a pre-defined set of topics (\textit{e.g.}, WiFi, pool, parking), we randomly sampled an equal number of hotel reviews per topic. Annotators labelled the sentiment of each aspect to \emph{positive, negative, mixed, neutral, not mentioned, wished for}. Besides the widely used sentiment labels, we introduced two custom labels, namely "not mentioned: a topic not being present in the text" and "wished for: an utterance with the intent of wishing for it". For example, "I wish the breakfast was free". We compute the F1-score \cite{van1979information} between prediction and ground truth for evaluation.

\subsection{Overall Sentiment Classification}
The purpose of this task is to identify overall sentiment within a text. The data mainly consists of hotel reviews and customer support conversations. The labels are \emph{positive, negative, neutral}. We use F1-score for evaluation.

\subsection{Aspect-based Text Segmentation}
The purpose of this task is to extract relevant segments within a text with respect to a given aspect of interest (e.g. "cleanliness", "service"). Using the same data as the ABSA task, human annotators were asked to label the start and the end of text span indices. We used BLEU-score \cite{papineni-etal-2002-bleu} to measure exact matches between the prediction and the ground truth.

\subsection{Topic Classification}
The purpose of this task is to identify the topics that a text references. We use the same hotel review dataset as the ABSA task. Given a review the model is expected to identify all relevant topics are mentioned from the following list: \emph{amenities, bar, beach, breakfast, cleanliness, comfort, location, noise, pool, price, restaurant, room, service, spa or gym, view, wifi}. We use F1-score for evaluation.

\subsection{HELM}
Holistic Evaluation of Language Models (HELM) \cite{liang2022holistic} defined an LLM-as-a-judge \cite{zheng2023judging} evaluation paradigm for natural language generation (NLG) such as faithfulness, relevance, compliance, and inclusiveness. We employed two types of evaluation setups: 1) scoring in the range of 1 to 5 scale, with 5 being the most-desired output  \cite{chiang2023closer,jing-etal-2025-scale}; and 2) binary classification with True or False labels \cite{wang2023chatgpt,liu2023g}.

\subsubsection{HELM Faithfulness}
The faithfulness metric measures the absence of hallucination or untrue facts derived from the LLM's own knowledge. A score of 5 means "highly-faithful" and a score of 1 means "high-unfaithful". Root Mean Square Error (RMSE)\cite{chai2014root} is used for evaluation.

\subsubsection{HELM Compliance}
Compliance provides a binary measure on whether the presence of content is compliant with legal policy. Example policies include toxic language, defamation, and infringements of intellectual property. F1-score is used for evaluation

\subsubsection{HELM Inclusiveness}
Inclusiveness is a binary metric which measures whether there exists the presence of hate speech, non-respectful language, harassment, and threats in the NLG. F1-score is used for evaluation.

\subsubsection{HELM Relevance}
Relevance measures how relevant the information is to the purpose of the task, for example : how relevant is the generated text given the task of review summarisation, on a 1-5 scale. Each use case has its individual measure of relevance. The rubrics change per use case. RMSE is used for evaluation.

\subsection{Review Moderation}
This task aims to identify whether a review is appropriate for publication. We evenly sampled hotel reviews and asked human annotators to label them as \emph{APPROVED or REJECTED}. We use F1-score for evaluation.

\subsection{Manager Response Moderation}
Similarly to review moderation, this task aims to classify property manager responses to customer reviews as \emph{APPROVED or REJECTED}.. Some common reasons for rejection include exposing sensitive personal information, being irrelevant to the review, or being overly promotional in order to bypass platform policies. We use F1-score for evaluation.

\subsection{Intent Prediction}
This task aims to determine the intent behind a customer's utterance with a customer service agent. We instructed human annotators to label each utterance as one of the following intents: \emph{book, cancel, change, contest\_payment, feedback, help, retrieve, small\_talk, unknown\_scenario}. We use F1-score for evaluation.

\subsection{Review Summarisation}
Using the same hotel review data as the ABSA task, we grouped the reviews by property and topic. Human annotators were instructed to summarise reviews within the same group into a short summary, which is less than 100 characters. We used METEOR-score \cite{banerjee-lavie-2005-meteor} for evaluation, which is robust to paraphrasing and synonyms.

\begin{figure*}[t]
    \centering
    \includegraphics[width=\linewidth]{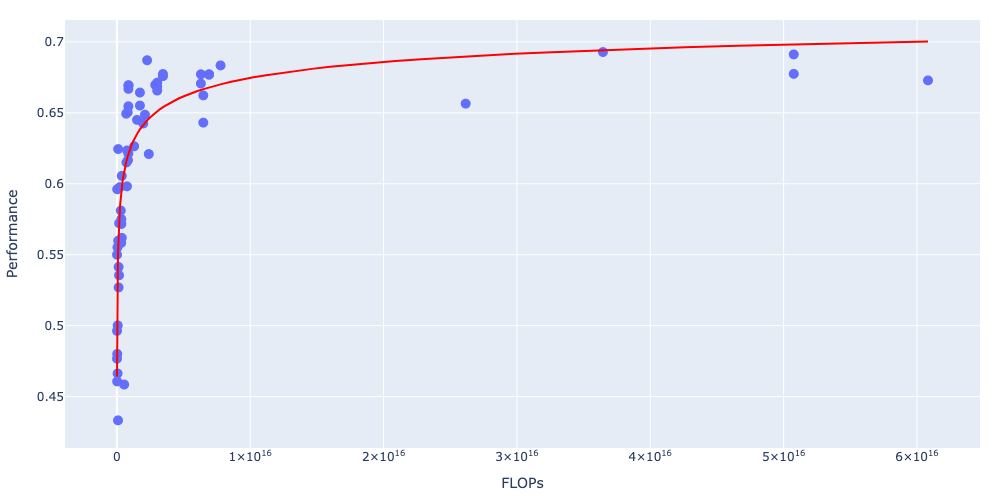}
    \caption{Performance $P_m$ against training compute $FLOPs$. While performance generally improves with scale, there are significant diminishing returns past $0.5*10^{16}$.}
    \label{fig:flops_scaling}
\end{figure*}

\subsection{Translation}
We split translation into two sub-tasks: 1) translate from English to a target language; and 2) translate from a target language to English. To obtain the annotations, we randomly sampled hotel reviews from non-English languages and instructed human annotators to translate target texts into English. METEOR-score was used for evaluation.

\section{Benchmark Setup}

We leveraged prompt engineering by treating each task as a chat completion problem. We used the OpenAI chat completion template via Python SDK to ensure response accuracy. For each task, we designed a few-shot prompt template and used the same template consistently across all LLMs. Each template consists of 1) the description of the task, 2) step-by-step instructions to solve the specific problem, and 3) Zero-shot or few-shot question/answer pairs. For better comparison, we limited our scope to instruct-tuned models only, and that a temperature of 0 and top-p of 1 were applied to obtain deterministic outputs. For models with reasoning capabilities, we adopted the default settings from their model cards. Additionally, to serve open-source LLMs, we utilized vLLM framework and for proprietary LLMs such as GPT-4o, we leveraged official APIs through a secure proxy. We focused on two evaluations: 1) overall performance comparison with $67$ models across open and closed source LLMs, and 2) the effect of reasoning of 8 hybrid-reasoning models from the Qwen 3 family.

\subsection{Scaled performance across metrics}
In order to ensure fair comparison of the models across various tasks, each with a different evaluation metric, we devised an overall performance metric $P_m$. Firstly, we calculated $P_{t,m}$ which denotes the scaled performance with respect to each task as the following, 
\begin{equation*}
    \resizebox{\columnwidth}{!}{$
    P_{t,m} = 
    \begin{cases}
        1 - \dfrac{\text{clamp}(x_{t,m}, a_t, b_t) - a_t}{b_t - a_t}, & \text{if } \text{invert}_t = \text{True} \\
        \\
        \dfrac{\text{clamp}(x_{t,m}, a_t, b_t) - a_t}{b_t - a_t}, & \text{if } \text{invert}_t = \text{False}
    \end{cases}
    $}
\end{equation*}
where $x_{t,m}$ is the original score for model $m$ on task $t$, $a_t$ is the theoretical minimum bound for task $t$, $b_t$ is the theoretical maximum bound for task $t$, $\text{invert}_t$ is an indicator if lower is better for task $t$, and $\text{clamp}(x, a, b) = \min(\max(x, a), b)$ is the clamping function to bound the score within $[a, b]$.

Then, we can calculate the overall performance $P_m$ by averaging over all tasks as, 
\begin{equation*}
    P_m = \frac{1}{N_m} \sum_{t} P_{t,m}
\end{equation*}
where $P_m$ represents the overall performance of model $m$, and $N_m$ represents the total number of valid (included) tasks for model $m$.

\begin{figure*}[t]
    \centering
    \includegraphics[width=\linewidth]{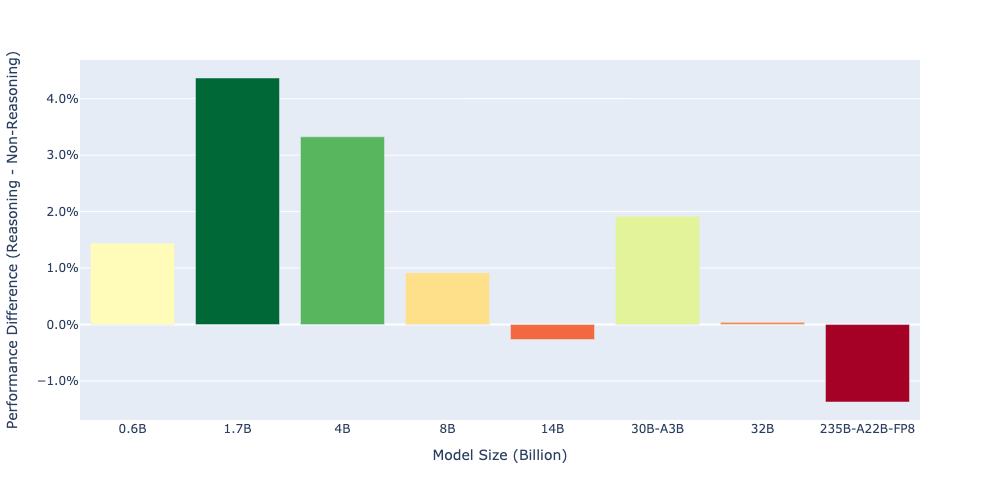}
    \caption{The effect of enabling reasoning across the Qwen3 model family. While smaller models show some improvements, the same does not apply to large models - performance degradation can be seen for the 235B model.}
    \label{fig:reasoning_vs_non_reasoning}
\end{figure*}

\section{Results and Analysis}
\label{sec:analysis}

\subsection{Scaling Laws}
We are interested in learning the correlation between model performance and its scale. Using the performance $P_m$ calculated from above and the compute approximation $FLOPs$ following \citeauthor{kaplan2020scalinglawsneurallanguage} (\citeyear{kaplan2020scalinglawsneurallanguage}), a model's scale can be defined as the training compute in floating-point operations per second (FLOPs),
\begin{equation*}
    FLOPs \approx 6NT
\end{equation*}
where $T$ is the number of training tokens and N is the number of model parameters

As illustrated in Figure \ref{fig:flops_scaling}, we can observe a positive correlation between model performance and scale across all tasks. However, the scaling progression rapidly diminishes as FLOPs increases. Models trained with compute budgets larger than $\approx 0.5*10^{16} $ FLOPs exhibits slow non-linear improvements, suggesting that unseen domain adaption remains a challenge even for larger models with higher generalizability. This result suggests that for domain adoption of out-of-the-box LLMs, lightweight models can sometimes serve as better options given the trade off between performance, compute costs, and inference latencies.

\subsection{Effects of Reasoning}
Recent work on unified training frameworks with toggle reasoning provided an opportunity to study its impact on the same model with in a model family. We tested the Qwen3 \cite{yang2025qwen3technicalreport} family models to measure the performance difference with and without enabling internal thinking.

Figure \ref{fig:reasoning_vs_non_reasoning} illustrates the performance differences for the model of the same size. As the model parameter increases, judging with reasoning does not always outperform that without reasoning. In addition, reasoning provides little to zero performance improvements on bigger models, and in some cases slightly degrades performance, which is counter-intuitive.

\subsection{Performance Variance}
\begin{figure*}[t!]
    \centering
    \includegraphics[width=\linewidth]{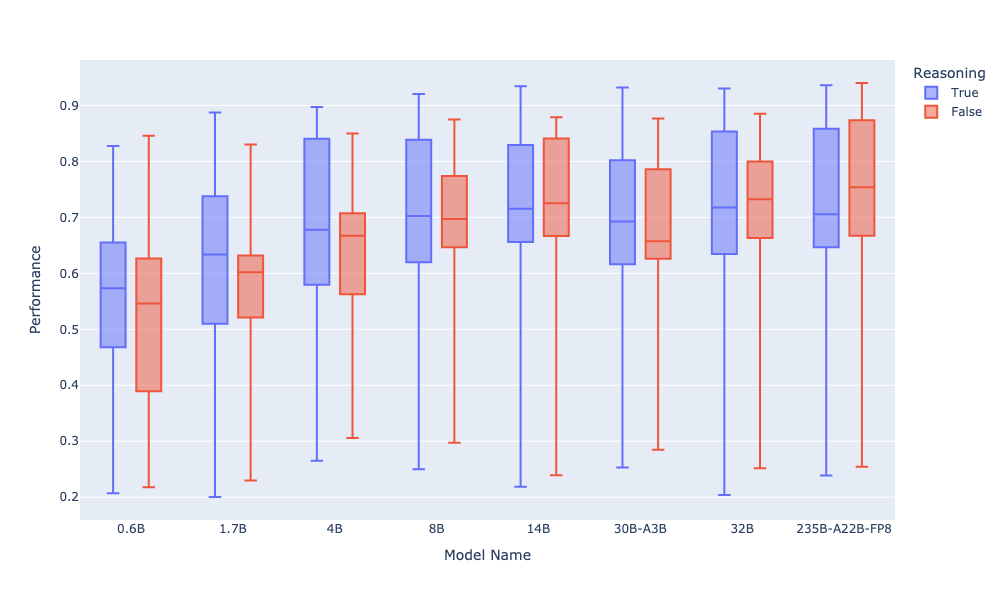}
    \caption{Variance of model performance across tasks across the Qwen 3 family. While the performance overall increases with model size, the spread of the performance does not follow the same pattern. This indicates that performance is very task dependent and no one model is the best at every task.}
    \label{fig:performance_variance}
\end{figure*}

Figure \ref{fig:performance_variance} demonstrates the variance in performance across the Qwen3 family models. We observe that larger models are on average better than their smaller counterparts when pre-trained on the same dataset. However, improvements in performance consistency are insignificant. We also notice that, while enabling reasoning improves the average performance of the smaller models, it can introduce more fluctuations in the model scores across tasks. However, this behaviour is less prevalent in larger models. Our assumption is that larger models store abundant internal knowledge with greater context retrieval ability. When a task is complex enough to overwhelm the model's natural retrieval capability, adding reasoning helps improve the recall. Yet reasoning does not provide the model with the ability to gain new knowledge, and thus the performance caps at a certain level.

\section{Related Works}

\textbf{Scaling laws.} \citeauthor{kaplan2020scalinglawsneurallanguage} (\citeyear{kaplan2020scalinglawsneurallanguage}) and \citeauthor{hoffmann2022trainingcomputeoptimallargelanguage} (\citeyear{hoffmann2022trainingcomputeoptimallargelanguage}) presented scaling laws for LLMs of varying sizes to understand the relationship between scaling model size and training data. They compared test loss and some general benchmark performance such as MMLU \cite{hendrycks2020measuring} and Big-Bench \cite{srivastava2023imitationgamequantifyingextrapolating}.

\textbf{Domain Specific Benchmarking. } Within the travel sector, TravelPlanner \cite{xie2024travelplanner} introduced a benchmark for evaluating LLMs on tool use and planning with synthetically generated queries. They compared both open source and closed source LLMs and showed that even the best performing model GPT4 has a success rate of only 0.6\%. ChinaTravel \cite{shao2025chinatravelopenendedbenchmarklanguage} extends this work by using open-ended human written queries for the Chinese travel market, also showing similar results with Deepseek V3 being the best model at 5\% pass rate, the authors note this may be because Deepseek has more context into chinese data. SemEval-2014 Task 4 \cite{cca5dae19d614c7c92c9acabc0428ce6} introduced an aspect-based sentiment analysis dataset on restaurant and laptop reviews, which has been widely used for benchmarking ABSA.

\textbf{Effects of reasoning. }Research such as \citeauthor{deepseekai2025deepseekr1incentivizingreasoningcapability} (\citeyear{deepseekai2025deepseekr1incentivizingreasoningcapability}) and \citeauthor{openai2024openaio1card} (\citeyear{openai2024openaio1card}) which train models with reinforcement learning to enable native reasoning in LLMs show prominent improvements in model performance through test-time compute scaling. We extend this research and compare new hybrid-thinking models proposed by \citeauthor{yang2025qwen3technicalreport} (\citeyear{yang2025qwen3technicalreport}) and show the effect of enabling reasoning traces. All prior research shows notable performance increases by introducing native reasoning.

\section{Conclusion}
We presented an in-depth evaluation of LLM-as-a-judge on a series of annotated datasets in the low-resource travel domain. We expanded the task variation from the previously prominent opinion mining to 7 common NLP tasks. Our results indicate that out-of-the-box LLMs, despite the number of parameters and the amount of training tokens, reach performance bottlenecks over domain-specific tasks. Furthermore, internal reasoning tends to have a more significant performance boost on smaller LLMs over larger models. Lastly, the scoring fluctuation over the Qwen-family models indicates that although reasoning helps with retrieval, there remains a gap between human annotation and model judgement on unseen knowledge.

\section{Limitations}
\begin{itemize}[nolistsep]
    \item Due to resource limitations, some of the models (>200b) tested were only tested in their quantised FP8/INT4 versions. This may have caused a slight degredation in performance for these models at the higher end of the FLOPs budget in \ref{fig:flops_scaling} however not enough to change the analysis drawn from it.
    \item We use a single prompt across all models, while this may not be optimal since different models might perform better with modified prompts we chose to keep the prompt identical as a measure to test ease of use. Future work could go into measuring how much the prompt matters in downstream performance.
    \item The internal company proxy service we used for closed-source models has inbuilt guardrails that protect against accidental misuse. However this caused some issues with dealing with some tasks such as review moderation as the harmful content filter was being triggered and caused the service to fail in providing a response. While this is testing whether the model can successfully filter out harmful content the proxy service would not allow it. So we had to remove the failing rows in our testing for those models.
    \item For closed-source models, we are not able to find details on training data size, quantisation and parameter size which meant we could not include those in the FLOPs scaling figures.
\end{itemize}


\bibliography{references_bib}

\end{document}